\documentclass[11pt,a4paper]{article}
\usepackage[hyperref]{emnlp2020}
\usepackage{graphicx}
\usepackage{amsmath, amssymb}
\usepackage{times}
\usepackage{latexsym}

\usepackage[ruled,vlined]{algorithm2e}
\usepackage{multirow}
\usepackage{caption}
\usepackage{subcaption}
\usepackage{diagbox}
\usepackage{amsfonts}
\usepackage[inline]{enumitem}
\usepackage{booktabs,dcolumn}
\usepackage{booktabs}
\usepackage{comment}

\usepackage{microtype}

\aclfinalcopy 

\title{Multimodal Speech Recognition with Unstructured Audio Masking}

\author{Tejas Srinivasan \\
  Language Technologies Institute \\
  Carnegie Mellon University \\
  \texttt{tsriniva@andrew.cmu.edu}\\\And
  Ramon Sanabria \\
  CSTR, ILCC \\
  University of Edinburgh \\
  \texttt{r.sanabria@ed.ac.uk}\\\AND
  Florian Metze\\
  Language Technologies Institute \\
  Carnegie Mellon University \\
  \texttt{fmetze@andrew.cmu.edu}\\\And
  Desmond Elliott\\
  Department of Computer Science\\
  University of Copenhagen \\
  \texttt{de@di.ku.dk}
   \\}

\date{}

\begin{document}
\maketitle

\begin{abstract}



Visual context has been shown to be useful for automatic speech recognition (ASR) systems when the speech signal is noisy or corrupted.
Previous work, however, has only demonstrated the utility of visual context in an unrealistic setting, where a fixed set of words are systematically masked in the audio.
In this paper, we simulate a more realistic masking scenario during model training, called RandWordMask, where the masking can occur for any word segment.
Our experiments on the Flickr 8K Audio Captions Corpus show that multimodal ASR can generalize to recover different types of masked words in this unstructured masking setting. Moreover, our analysis shows that our models are capable of attending to the visual signal when the audio signal is corrupted.
These results show that multimodal ASR systems can leverage the visual signal in more generalized noisy scenarios.
\end{abstract}

\section{Introduction}
\label{sec:intro}


 


Jointly modelling linguistic and visual signals is beneficial for several language processing tasks, such as machine translation \cite{sulubacak2019multimodal}, visual question-answering (VQA) \cite{antol2015vqa}, summarization~\cite{palaskar2019multimodal} and automatic speech recognition (ASR) \cite{palaskar2018end,sanabria18how2}. However, it is unclear exactly how the visual signals are useful for these tasks. For example, in VQA, it has been observed that models can ignore the visual context and instead rely on linguistic biases in the dataset~\cite{ramakrishnan2018overcoming, grand2019adversarial}; in machine translation, it has been shown that some models are not affected by incorrect visual signals~\cite{desmond}; and in multimodal ASR, the visual signals were shown to act as a regularizer instead of useful disambiguating context~\cite{Caglayan2018multimodal}. Given these uncertainties, there is a need to clarify the circumstances in which visual signals are useful.

    

Previous work in multimodal machine translation~\cite{caglayan2019probing} and ASR~\cite{srinivasan2020looking} shows that the visual signal is useful when the linguistic signal is degraded by dropping the input. In this setting, multimodal models leverage the visual signals to recover the missing language information. The results in~\cite{srinivasan2020looking} are a promising start towards \textit{verifiably useful} multimodality for robust speech recognition. However, the experiments were conducted with structured noise that focused on a predetermined set of groundable entities (\textit{i.e.}, nouns and places). In real world scenarios, however, noise occurs in a more unstructured manner. Therefore, it is important that multimodal models can use the visual signal in a wider variety of situations. 


\begin{figure}[t]
\centering
\includegraphics[width=\columnwidth]{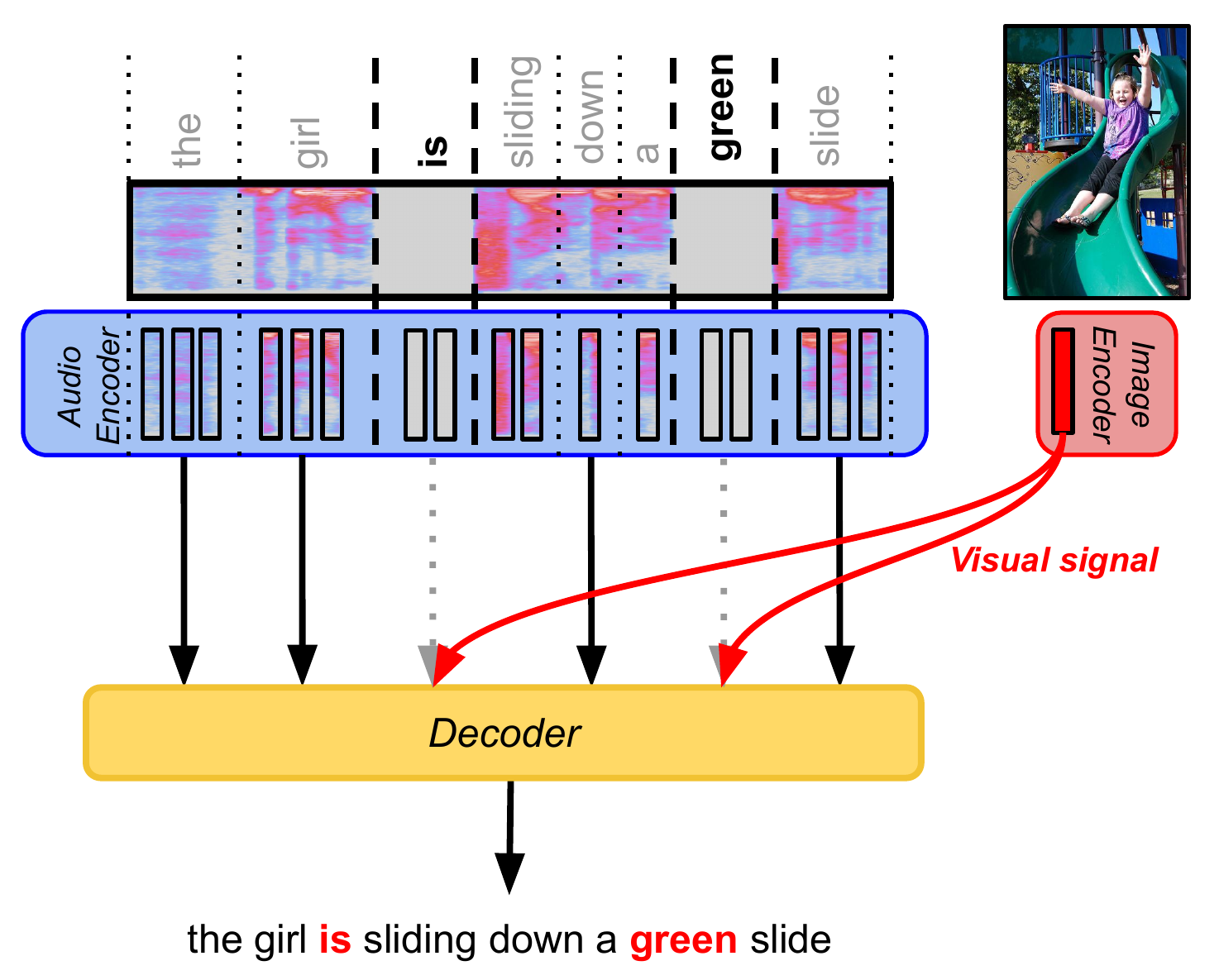}
\caption{We propose to train multimodal speech recognition models while randomly masking different types of words in the speech signal. The model learns to use the visual signal to correctly predict the masked words.}
\label{fig:main-fig}
\end{figure}

\begin{figure*}
    \centering
    \includegraphics[width=\textwidth]{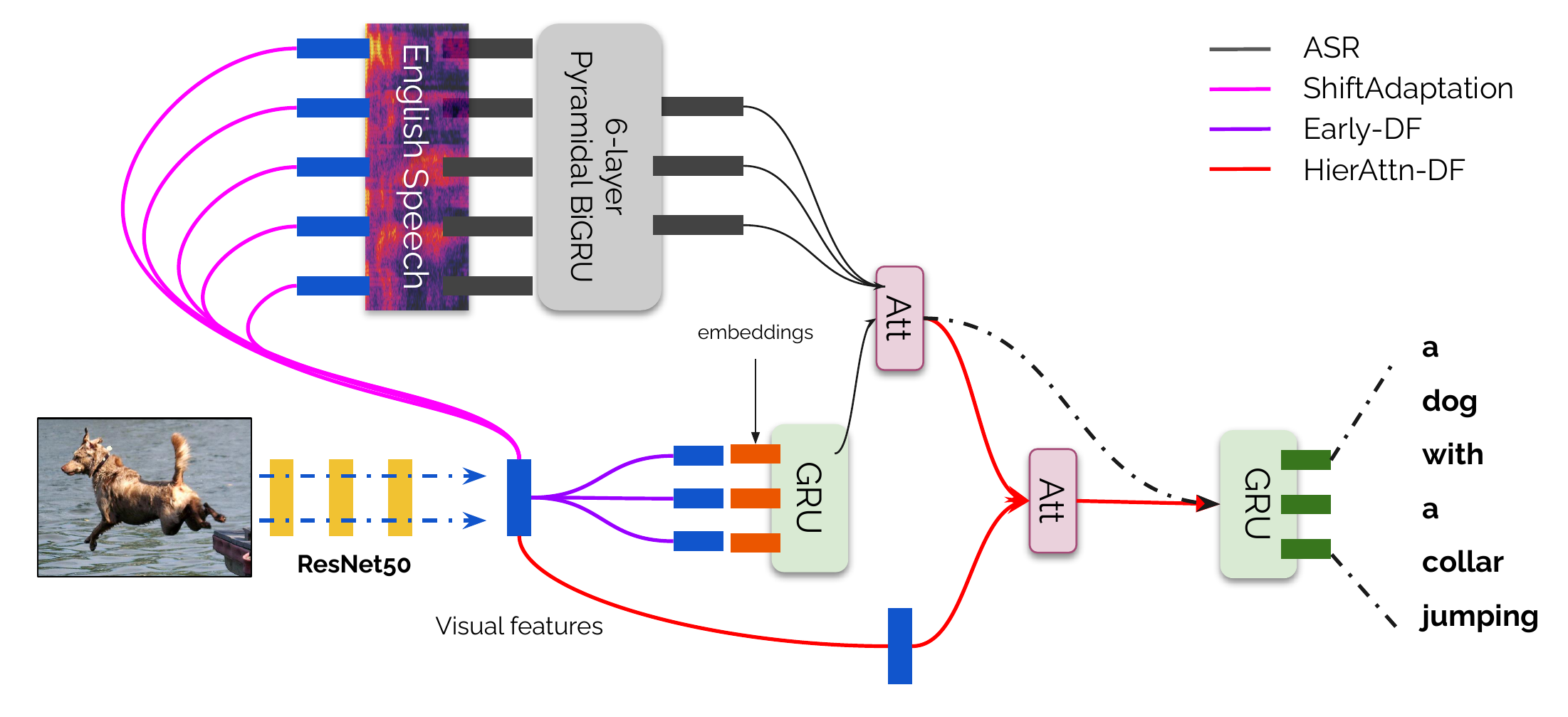}
    \caption{Our unimodal ASR model, along with several of our fusion methods for integrating a visual context vector (in blue) into the ASR model. The two fusion methods not displayed above, Weighted-DF and Middle-DF, were constructed similar to Early-DF and HierAttn-DF respectively}
    \label{fig:model}
\end{figure*}

In this work, we study multimodal ASR in more realistic noisy scenarios. We follow the methodology from~\cite{srinivasan2020looking} but we randomly mask words in an unstructured manner in the audio signal (we refer to this as RandWordMask). This is in contrast to the structured masking in~\cite{srinivasan2020looking}, where the masked audio corresponds to only entities (which we refer to as EntityMask). The example in Figure \ref{fig:main-fig} shows that RandWordMask can mask any words in the audio signal, whereas EntityMask would only mask entities like ``girl'' and ``slide''. We apply masking both during training and testing. 




The main contributions of this work are:
\begin{itemize}
    \item We simulate a more realistic masking scenario, called RandWordMask\footnote{
We note that RandWordMask is different from robust ASR~\cite{barker2018fifth} scenarios, where the whole signal is corrupted with stationary noise.
}, during training and testing of our ASR models (Section~\ref{sec:method}). 
    \item We propose several multimodal models (Section~\ref{subsec:mm-asr}), and show that training with RandWordMask improves their ability to recover masked words (Section~\ref{sec:results}).
    \item We show that our multimodal ASR models are right for the right reasons through several quantitative analyses (Section~\ref{subsec:hier-attn}, \ref{subsec:randmask-utility}, \ref{subsec:cong}). 
\end{itemize}


The results show that visual signals improve speech recognition in this more difficult, unstructured setting where random words are masked. Our models are not only able to recover masked entities, but they also recover words from other syntactic categories, \textit{e.g.}, adjectives, cardinals, and verbs. Furthermore, our analysis shows that our models when trained using RandWordMask attend to the visual signal when the audio signal is unavailable. This confirms that the visual context can be leveraged when the primary audio signal is masked.



\section{Methodology}
\label{sec:method}

In this section, we describe the different ASR models and our technique for simulating unstructured audio masking.

\subsection{Unimodal ASR Model}

Our unimodal ASR model is a word-level~\cite{palaskar2018acoustic} sequence-to-sequence model with attention~\cite{bahdanau2016end, chan2016listen}, identical to the model used in~\cite{srinivasan2020looking}. The encoder ($\mathbf{E}$) consists of 6 bidirectional LSTM layers \cite{schuster1997bidirectional,hochreiter1997long} with temporal sub-sampling~\cite{chan2016listen} in the middle two layers. The decoder is a two-layer conditional gated-recurrent-unit~\cite{cho2014learning} which computes attention over the encoder states $\mathbf{E}$.
\begin{align}
    \mathbf{h_t^{dec1}} & = \text{GRU$_1$}(\mathbf{y_{t-1}}, \mathbf{h_{t-1}^{dec1}}) \\
    \mathbf{z_t} & = \text{Attention}(\mathbf{E}, \mathbf{h_t^{dec1}}) \label{eqn:zt}\\
    \mathbf{h_t^{dec2}} & = \text{GRU$_2$}(\mathbf{z_t}, \mathbf{h_{t-1}^{dec2}}) \label{eqn:dec2}
\end{align}

\subsection{Multimodal ASR Models}
\label{subsec:mm-asr}

We explore several fusion methods to integrate a visual feature vector $\mathbf{v}$ into the unimodal ASR model.

\textbf{Encoder Feature Fusion:}
We use a visual adaptation method similar to \cite{Caglayan2018multimodal}, which we call \textbf{Shift Adaptation}. The visual feature vector $\mathbf{v}$ is projected down to the speech feature dimension; the resulting ``shift vector'' $\mathbf{s}$ is then added to the input speech features at all timesteps.
\begin{align}
    \mathbf{s} & = \mathbf{W_v} \mathbf{f} + \mathbf{b} \\
    \mathbf{x_t} & = \mathbf{x_t} + \mathbf{s} \quad \quad \forall \mathbf{t} \in \{1,...,T\}
\end{align}

\textbf{Decoder Feature Fusion:}
Instead of integrating the visual features into the encoder, we can integrate them in the decoder. We hypothesize that this will bias the ASR's language modelling capacity. Anastasopoulos \textit{et al.}\cite{anastasopoulos2019neural} explore several strategies for incorporating visual features into an LSTM language model. We employ similar fusion methods in our decoder.
\begin{enumerate}
    \item \textbf{Early Decoder Fusion (Early-DF):} At each timestep, we concatenate $\mathbf{v}$ to the input embedding $\mathbf{y_t}$, which is then projected down to the embedding dimension.
    \begin{align}
        \mathbf{y_t} &= \mathbf{W_{proj}} \mathbf{[y_t; v]}
    \end{align}
    \item \textbf{Weighted Early Decoder Fusion (Weighted-DF):} We calculate a timestep-dependent weighted scalar between the input embedding $\mathbf{y_t}$ and the embedded visual features $\mathbf{v}$ (Eqn. \ref{eqn:scalar}), which scales the contribution of the visual features in the concatenated input (Eqn. \ref{eqn:scaled}):
    \begin{align}
        \lambda &= \sigma(\mathbf{y_t} \cdot \mathbf{v}) \label{eqn:scalar}\\
        \mathbf{y_t} &= \mathbf{W_{proj}} [\mathbf{y_t}; \lambda \mathbf{v}] \label{eqn:scaled}
    \end{align}
    \item \textbf{Middle Decoder Fusion (Middle-DF):} In this approach, fusion occurs between the GRU layers at $\mathbf{z_t}$ (Eqn. \ref{eqn:zt}), which is the input to the 2nd decoder layer:
    \begin{align}
        \mathbf{z_t} &= \mathbf{W_{proj}} [\mathbf{z_t; v}]
    \end{align}
    \item \textbf{Hierarchical Attention over Features (HierAttn-DF):} In this approach, we add a hierarchical attention layer~\cite{libovicky2017attention} that attends between the encoder context vector $\mathbf{z_t}$ (Eqn. \ref{eqn:zt}) and the visual feature vector $\mathbf{v}$. The hierarchical context vector $\mathbf{z_t^{hier}}$ is the input to the second decoder layer (Eqn. \ref{eqn:dec2}):
    \begin{align}
        \mathbf{z_t^{hier}} = \text{Attention}(\mathbf{\{z_t, v\}}, \mathbf{h_t^{dec1}})
    \end{align}
    By conditioning the hierarchical attention on the output of the first decoder layer, the attention layer learns to decide which of the audio and visual modalities is more important for decoding at a given timestep.
\end{enumerate}

\subsection{Unstructured Masked Audio: RandWordMask}
\label{subsec:randwordmask}
We simulate a degradation of the audio signal by randomly masking words in the audio with silence. This approach differs from~\cite{srinivasan2020looking}, where they masked a fixed set of words corresponding to entities, i.e., nouns and places. Figure~\ref{fig:main-fig} shows an example of an audio spectrogram with \textbf{RandWordMask}. The intuition behind random word masking, as opposed to entity-based word masking, is that noise in the audio signals is unlikely to systematically occur when someone is speaking about an entity. Our multimodal ASR models need to be responsive to audio that drops outside systematically expected regions.

In real-world settings, the rate at which the speech is masked (unavailable) is highly variable. Therefore, we train the models with an augmented version of the dataset: for each audio utterance, we create four masked audio samples, where words are masked with 0\%, 20\%, 40\% and 60\% probability. Note that the text transcript ($\mathbf{y_{1 \ldots N}}$) and image modality ($\mathbf{v}$) remain intact. This approach to augmenting the dataset will result in models that can adapt to different amounts of corruption in the audio signal during evaluation.


\section{Experimental Setup}
\label{sec:experiments}




\subsection{Dataset}
We perform experiments on the Flickr 8K Audio Caption Corpus~\cite{harwath2015deep},
which contains 40,000 spoken captions (total 65 hours of speech)
corresponding to 8,000 natural images from the Flickr8K
dataset~\cite{DBLP:conf/ijcai/HodoshYH15}. The augmented dataset that we use for training and testing (as described in Section~\ref{subsec:randwordmask}) consists of 160,000 spoken captions.

In addition, we use the SpeechCOCO dataset~\cite{Havard2017} for pretraining. SpeechCOCO contains over 600 hours of \textit{synthesised} speech paired with images.

\subsection{Implementation Details}

\subsubsection{Audio Features}
We extract 43-dimensional filter bank features in an identical manner to~\cite{srinivasan2020looking}. In order to mask the audio, we first extract word-audio alignments from a pre-trained GMM-HMM model and expand the start and end timing marks by 25\% of the segment duration to account for misalignments. We mask words in the audio by replacing word segments with 0.5 seconds silence.

\begin{table*}[t]
\centering
\begin{subtable}{\textwidth}
\begin{tabular}{lccccccccccc}
\toprule
\multicolumn{1}{c}{} & \phantom{a} & \multicolumn{4}{c}{ $\uparrow$ Recovery Rate (\%)} & \phantom{a} & \multicolumn{5}{c}{ $\downarrow$  Word Error Rate (\%)}\\
\cmidrule(lr){3-6} \cmidrule(lr){8-12}
 Masking Perc. & \phantom{a} & Aug. & 20\% & 40 & 60\% & \phantom{a} & Aug. & 0\% (Clean) & 20\% & 40\% & 60\% \\
\midrule
Unimodal               & \phantom{a} & 29.3              & 36.5               & 30.9               & 24.7               \phantom{a} & & 34.0                 & 13.7                  & 26.3           & 40.7               & 57.1               \\        
\midrule
ShiftAdapt             & \phantom{a} & 29.3              & 36.5               & 31.3               & 25.1               \phantom{a} & & 34.0                 & 13.5                  & 25.9           & 40.5               & 57.1               \\        
Early-DF               & \phantom{a} & 32.0              & 38.2               & 33.2               & 28.7               \phantom{a} & & 33.3                 & 13.7                  & 25.9           & 39.7               & 55.3               \\       
Weighted-DF            & \phantom{a} & 33.0              & 38.8               & 34.5               & 29.6               \phantom{a} & & \bf{32.6}            & \bf{13.4}             & \bf{25.5}      & \bf{38.9}          & \bf{53.9}          \\     
Middle-DF              & \phantom{a} & 32.4              & 37.9               & 34.1               & 29.7               \phantom{a} & & 34.1                 & 14.6                  & 26.9           & 40.3               & 55.3               \\       
\midrule
HierAttn-DF            & \phantom{a} & \bf{33.5}         & \bf{40.3}          & \bf{35.2}          & \bf{30.1}          \phantom{a} & &  33.2                & 13.9                  & 25.9           & 39.3               & 54.7               \\
\bottomrule
\end{tabular}
\centering
\caption{Recovery Rate (RR) and Word Error Rate (WER) of the ASR models on the FACC development set.}
\label{tab:main-results}
\end{subtable}
\begin{subtable}{\textwidth}
\centering
\vspace{1em}
\begin{tabular}{ccccccccc}
\toprule
 Metric                        & Model         & Nouns   & Places  & Adj. & Colors  & Verbs   & Adverbs & Cardinals \\ 
\cmidrule(lr){1-1} \cmidrule(lr){2-2} \cmidrule(lr){3-9}
\multirow{2}{*}{RR (\%)}         & Unimodal    & 37.2 & 28.0 & 26.0    & 26.6 & 26.0 & 30.4 & 56.7   \\ 
& HierAttn-DF  & 47.9 & 40.0 & 29.7    & 30.4 & 27.9 & 29.2 & 58.1   \\ \cmidrule(lr){1-1} \cmidrule(lr){2-2} \cmidrule(lr){3-9}
  Rel. $\Delta$ RR (\%) & - & 28.8 & 42.8 & 14.2 & 14.3 & 7.3 & -3.9 & 2.4 \\ 
\cmidrule(lr){1-1} \cmidrule(lr){2-2} \cmidrule(lr){3-9}
 G.R. (\%) & HierAttn-DF & 92.7 & 92.5 & 76.8    & 75.5 & 67.6 & 33.5 & 82.3   \\ \bottomrule
\end{tabular}
\caption{Comparison of Recovery Rates of unimodal and HierAttn-DF ASR on various syntactic and semantic word categories.}
\label{tab:hierattn-results}
\end{subtable}
\caption{Recovery Rate, Word Error Rate, and Grounding Rates for the proposed models on the FACC dataset.}
\end{table*}
\normalsize

\subsubsection{Visual Features}

We extract visual features from a ResNet-50 CNN~\cite{DBLP:conf/cvpr/HeZRS16} pre-trained on ImageNet. Specifically, we extract features from the 2048-dim average pooling layer, and project these to 256-dim through a learned linear layer: $\mathbf{v} = \mathbf{W}\cdot\text{CNN}(\mathbf{img})$

\subsubsection{Model Implementation}
We use the same model hyperparameters as in~\cite{srinivasan2020looking}. Models are trained using the \textit{nmtpytorch} framework~\cite{DBLP:journals/pbml/CaglayanGBABB17}. We first pre-train our models for 25,000 minibatches on the SpeechCOCO dataset. This pre-training step, inspired by~\cite{ilharco2019large}, was crucial to ensure stable training of our models on the Flickr 8K dataset. 

\subsection{Evaluation Metrics}
Our model evaluation (Table~\ref{tab:main-results}) has been conducted on the development set of Flickr8k-Audio, while the rest of our analysis is conducted on the test set. We report \textbf{WER} for all our models. For datasets where words have been masked in the audio signal, we compute \textbf{Recovery Rate}~\cite{srinivasan2020looking}, which measures the percentage of masked words which have been correctly recovered in the transcription.

In addition, we can determine the contribution of the visual signal when decoding each word in the HierAttn-DF model. We do this by inspecting the weights of the audio and visual modalities in the hierarchical attention mechanism. We introduce a new metric to quantify this: \textbf{Grounding Rate (G.R.)}.
\begin{align*}
    \text{G.R.} = \frac{\# \text{recovered words where visual attn} > 0.5}{\# \text{correctly recovered masked words}}
\end{align*}
We choose 0.5 as the threshold since above this value, more attention was given to the visual modality than the audio. G.R. thus represents the percentage of recovered words where the model was focusing more on the visual context while decoding.

\section{Results and Analysis}
\label{sec:results}
In Table \ref{tab:main-results}, we summarize the performance of our unimodal ASR and proposed multimodal ASR models. Our development set is constructed similarly to our training set described in Section \ref{subsec:randwordmask}, consisting of samples with 0\%, 20\%, 40\% and 60\% of words masked. We examine performance on this Augmented dataset, as well as datasets at each individual masking level.

We see that the Decoder-Fusion (DF) multimodal models outperform unimodal ASR on both WER and RR. However, the best-performing models on both metrics differ: Weighted-DF achieves the lowest WER, with an improvement of 1.40\% on the augmented dataset. HierAttn-DF has the best Recovery Rate, with an absolute improvement of 4\% over the Unimodal model. These trends hold across all masking levels. Moreover, we observe that as the amount of masking in the audio signal increases, the WER and RR gains of our models increase. The ShiftAdapt model, which integrates the visual features with the speech encoder input, does not show any improvements over unimodal ASR. We observe that ShiftAdapt shows improvements when trained and tested on clean data, which aligns with the regularization signal previously observed in~\cite{Caglayan2018multimodal}.

The results in Table~\ref{tab:main-results} show that multimodality can recover words which were masked in an unstructured manner. We now turn our attention to analysing which types of words are recovered better. We conduct this analysis across seven categories: five syntactic (nouns, verbs, adjectives, adverbs and cardinals) and two semantic (places and colors).\footnote{Words for the syntactic categories were found by POS tagging the dataset and keeping the top 100 frequent words.} For each category, we create a new test set where we mask all word occurrences. We note that these categories are varying degrees of ``groundable'', which we define as how easily identifiable they are in the visual modality - the more groundable a category, the easier it is to identify words belonging to that category in the visual context. Nouns and places are the most groundable categories, while adjectives and colors are also frequently easy to identify in the image. Verbs and adverbs, however, are less groundable categories.

In Table~\ref{tab:hierattn-results}, we compare the Recovery Rate of the unimodal ASR and HierAttn-DF (the best multimodal model in terms of RR) on the different word types. We observe that on the groundable entities~\textit{i.e.}, nouns and places, there is a relative improvement of at least 25\% compared to the Unimodal model. Adjectives and colors, which are also groundable in the visual modality, are recovered around 14\% better than the Unimodal model. The relative RR improvement for verbs is around 7\%, whereas adverbs recovery is 4\% worse. These results show that visual context can recover words from a variety of categories, even though it is better at recovering entities, and struggles with words that are less groundable in the image.

\begin{table*}[t]
\centering
\begin{tabular}{c l c}

     \toprule
     Image & Reference Caption & Hierarchical Attention Map w/ Hypothesis Decoding \\ \midrule \\& & \vspace{-10 mm} \\
     \multirow{3}{*}{\includegraphics[width=2.5cm]{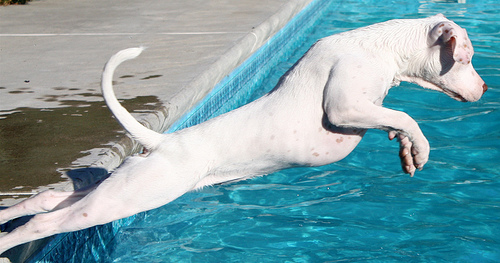}} & & \multirow{3}{*}{\includegraphics[width=7cm]{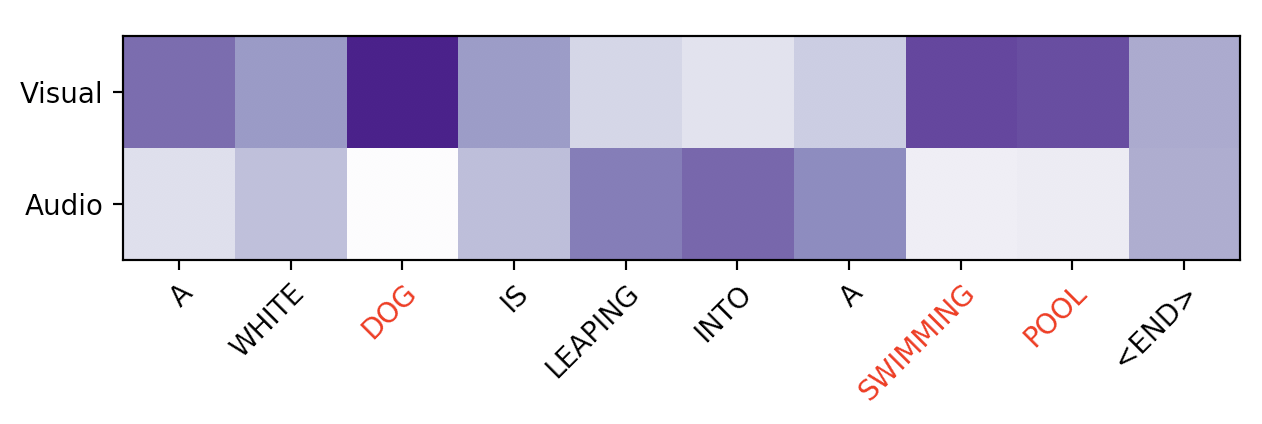}} \\
     & \begin{tabular}[c]{@{}l@{}}A white \textcolor{red}{dog} is\\ leaping into a\\ \textcolor{red}{swimming pool}\end{tabular} & \\
     &  & \vspace{-2mm}
     \\

     \midrule
     \multirow{3}{*}{\includegraphics[width=2cm]{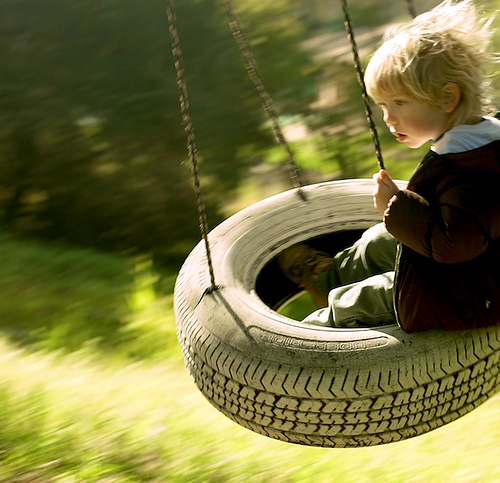}} & & \multirow{3}{*}{\includegraphics[width=7cm]{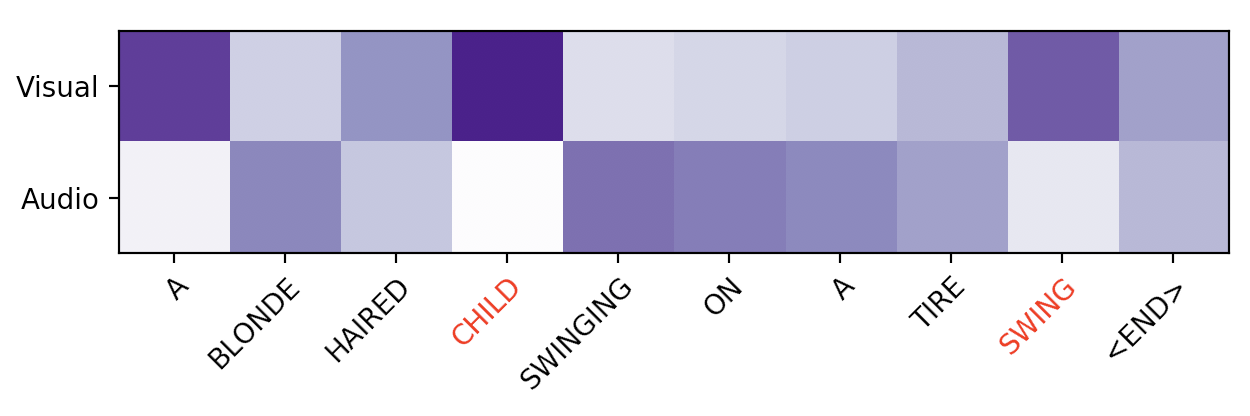}} \\
     & \begin{tabular}[c]{@{}l@{}}A blonde haired\\ \textcolor{red}{toddler} swinging on\\ a tire \textcolor{red}{swing}\end{tabular} & \\
     &  & \vspace{-7mm}
     \\
     \\
     \bottomrule
     
\end{tabular}
\caption{Examples of the HierAttn-DF model attending to the visual modality to recover \textcolor{red}{masked words} }
\label{tab:examples}
\end{table*}

\subsection{Hierarchical Attention Analysis}
\label{subsec:hier-attn}

In Table~\ref{tab:hierattn-results}, we also summarize the Grounding Rate of HierAttn-DF when recovering different types of words. We find that the most groundable words (nouns and places), have a Grounding Rate $>90\%$. This means that 90\% of the time the nouns/places were correctly recovered, the visual modality was being attended to. 
Adjectives and verbs, which are also groundable, have a grounding rate of $\approx 76\%$. These trends confirm that the model's improvements in masked word recovery are coming from using the visual signal.

\begin{figure}[t]
    \centering
    \includegraphics[width=\columnwidth]{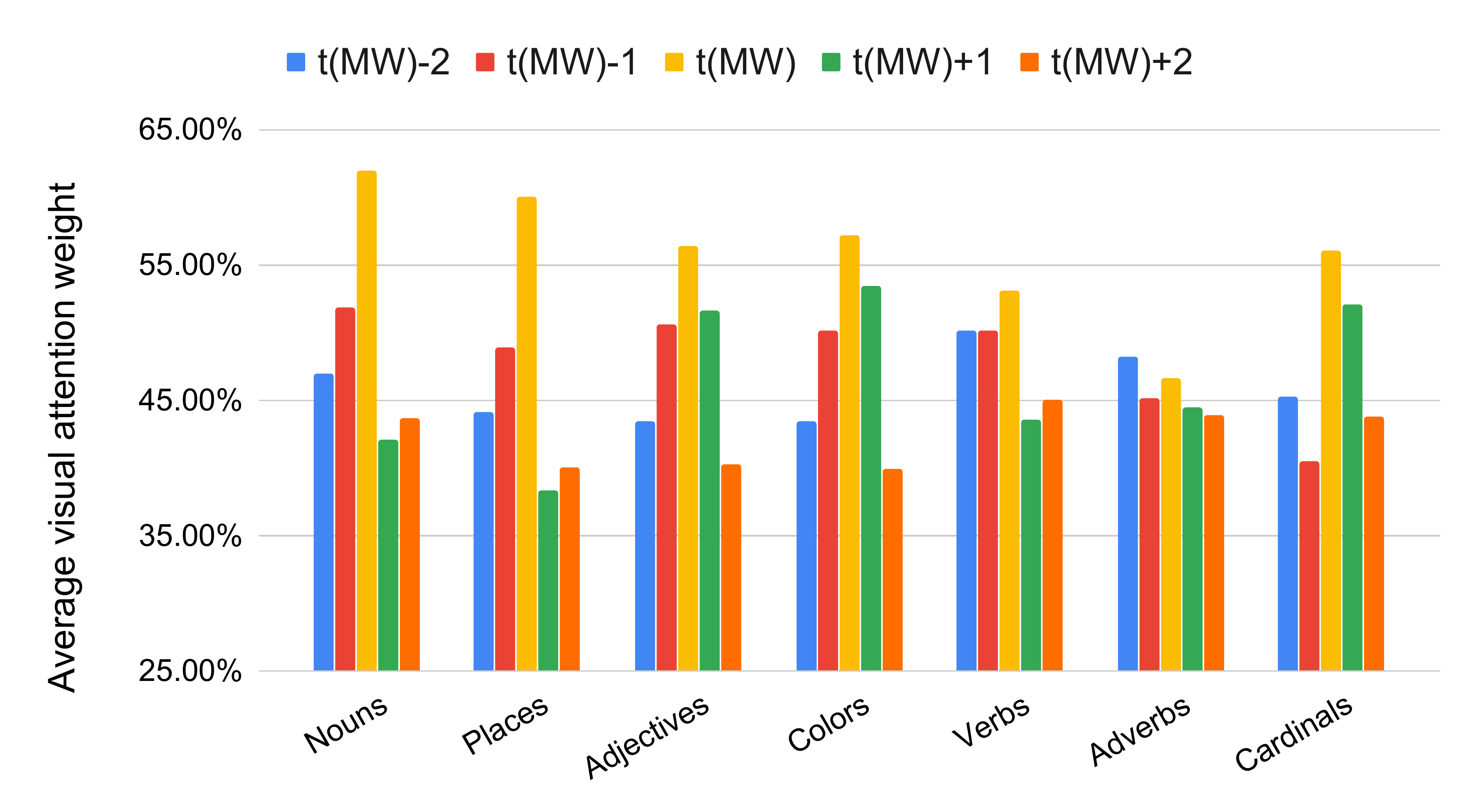}
    \caption{Average visual attention weight preceding and proceeding the onset of the masked word at timestep t(MW).}
    \label{fig:mw}
\end{figure}

In addition to calculating the Grounding Rate, we also check whether the model learns to ``look'' at the visual modality when it encounters a masked word. In Figure~\ref{fig:mw}, we plot the average visual attention weight at the masked word timestep, as well as the two preceding and proceeding timesteps. We see that the more groundable the word category, the more attention it learns to pay to the visual modality when the word is masked.

In Table~\ref{tab:examples}, we present some qualitative examples where we visualize how the attention to each modality evolves with time. We observe that the timesteps corresponding to masked words in the signal have significantly higher visual attention. We see that in the first example, all masked words are correctly recovered. In the second example, however, the model replaces the word \textit{toddler} with \textit{child}, which are semantically similar and visually identical, showing that the model knows what to recover in the image but does not always recover it in the correct form.

\subsection{Utility of RandWordMask Training}
\label{subsec:randmask-utility}

\begin{table}[b]
\centering
\resizebox{\columnwidth}{!}{%
\begin{tabular}{lccc}
\toprule
Masked Word & None & EntityMask & RandWordMask \\
\cmidrule(lr){1-1} \cmidrule(lr){2-4}
Nouns       & 4.3     & 59.1    & 47.9      \\ 
Places      & 2.4     & 43.1    & 40.0      \\ 
Adjectives  & 0.7     & 4.7     & 29.7      \\ 
Colors      & 1.3     & 3.4     & 30.3      \\ 
Verbs       & 0.7     & 11.9    & 27.9      \\ 
Adverbs     & 1.1     & 4.6     & 29.2      \\ 
Cardinals   & 3.5     & 4.3    & 58.1      \\ 
\bottomrule
\end{tabular}}
\caption{RR (\%) of different training schemes}
\label{tab:training}
\end{table}

We compare our RandWordMask training scheme with the EntityMask training mechanism from ~\cite{srinivasan2020looking}. In EntityMasking, only entities (nouns) are masked during training, and we hypothesize that this makes the model better at recovering entities but unable to generalize to other word types. Since RandWordMask training involves masking words at random, we expect the model should be able to generalize better to other words types. In Table \ref{tab:training}, we compare the performance of the HierAttn-DF model when trained with three different training mechanisms: \begin{enumerate*}[label=(\roman*)]
  \item None: no words are masked during training,
  \item EntityMask: top 100 frequent nouns are masked,
  \item RandWordMask.
\end{enumerate*}
As expected, when trained without masking in the training set (None), the model recovers almost none of the masked words. While EntityMasking shows strong performance on recovering nouns and places (which are closely related), it doesn't generalize to the other syntactic/semantic word types. RandWordMask results in slightly worse performance on noun recovery, but it generalizes much better to other word categories.

\subsection{Silence vs Whitenoise Masking}
Our results in Tables~\ref{tab:main-results} and \ref{tab:hierattn-results} are performed in the experimental setting where words are masked with silence. However, another masking strategy explored in~\cite{srinivasan2020looking} is white noise masking, where the masked word is replaced with white noise in the audio signal. \citep{srinivasan2020looking} had reported results in both masking scenarios, and noted that the improvements of the multimodal ASR model were similar in both scenarios. We further verify this by training unimodal and HierAttn-DF ASR models using RandWordMask, but with white noise masking instead of silence.

\begin{table}[t]
\centering
\begin{tabular}{ccc}
\toprule
\multicolumn{3}{c}{Silence Masking}    \\
\cmidrule{1-3}
Masking \% & Unimodal & HierAttn-DF \\
\cmidrule(lr){1-1} \cmidrule(lr){2-3}
20\%          & 36.6     & 40.6        \\
40\%          & 31.1     & 36.0        \\
60\%          & 25.7     & 31.3        \\
\toprule
\multicolumn{3}{c}{Whitenoise Masking} \\
\cmidrule{1-3}
Masking \% & Unimodal & HierAttn-DF \\
\cmidrule(lr){1-1} \cmidrule(lr){2-3}
20\%          & 33.1     & 37.4        \\
40\%          & 26.7     & 32.1        \\
60\%          & 21.5     & 28.1    \\
\bottomrule
\end{tabular}
\caption{RR (\%) of unimodal and HierAttn-DF ASR models when trained and tested on silence and white noise masked audio, at different masking levels }
\label{tab:whitenoise}
\end{table}

In Table~\ref{tab:whitenoise}, we report the Recovery Rates of both ASR models in both silence and white noise masking scenarios. We observe that while recovery is generally harder with white noise masking (evidenced by lower RR of both unimodal and multimodal ASR models), the HierAttn-DF model shows approximately the same absolute improvements in RR over the unimodal ASR. This indicates that the multimodal model can be applied to the more difficult white noise masking as well.

\subsection{Congruency Analysis}
\label{subsec:cong}

We perform a sanity check of our model by misaligning audio utterances and images while decoding the trained model~\cite{desmond}. This evaluation quantifies the sensitivity of the model towards the visual modality. A model that is sensitive to the visual context would perform significantly worse when presented with an unrelated (\textit{incongruent}) image during evaluation. Since the model has been trained to actively use the image, it is likely to extract incorrect information.
In Table~\ref{tab:incong}, we see that the HierAttn-DF model is substantially affected by the unrelated images (the recovery rate drops on average by 7\%). This verifies that our multimodal models are sensitive to the image modality.

\begin{table}[t]
\centering
\begin{tabular}{ccc}
\toprule
Masking \% & Congruent & Incongruent \\ \cmidrule(lr){1-1} \cmidrule(lr){2-3}
20\%                                       & 40.6                         & 29.3                           \\ 
40\%                                       & 36.0                         & 24.7                           \\ 
60\%                                       & 31.3                         & 20.2                           \\ 
\bottomrule
\end{tabular}
\caption{Recovery Rates (\%) for the HierAttn-DF model when provided with correct (congruent) and misaligned (incongruent) image}
\label{tab:incong}
\end{table}




\section{Conclusions}
We show that visual signals improve multimodal speech recognition when the audio signal is subject to unstructured masking. RandWordMask simulates a wider range of noisy scenarios by masking different types of words in the audio signal during training and evaluation, as opposed to previous work that only masked groundable entities~\cite{srinivasan2020looking}. Future work involves developing new models that attend over visual features extracted from object proposals, which provide better visual signals.

\section*{Acknowledgments}
This work used the computational resources of the PSC Bridges cluster at Extreme Science and Engineering Discovery Environment (XSEDE)~\cite{6866038}
\bibliographystyle{acl_natbib}
\bibliography{anthology,emnlp2020}

\appendix

\end{document}